\documentclass{article} 
\usepackage{iclr2020_conference,times}

\usepackage{amsmath,amsfonts,bm}

\def\eqref#1{equation~\ref{#1}}

\def\1{\bm{1}}

\DeclareMathAlphabet{\mathsfit}{\encodingdefault}{\sfdefault}{m}{sl}
\SetMathAlphabet{\mathsfit}{bold}{\encodingdefault}{\sfdefault}{bx}{n}

\usepackage{hyperref}
\usepackage{url}

\title{Towards Supervised and Unsupervised Neural Machine Translation Baselines for Nigerian Pidgin}

\author{%
  Orevaoghene Ahia\thanks{Equal Contribution}\\
  InstaDeep\\
  \texttt{o.ahia@instadeep.com} \\
 \And
   Kelechi Ogueji\footnotemark[1]  \\
   InstaDeep \\
   \texttt{k.ogueji@instadeep.com} \\
}

\iclrfinalcopy 
\begin{document}

\maketitle

\begin{abstract}
Nigerian Pidgin is arguably the most widely spoken language in Nigeria. Variants of this language are also spoken across West and Central Africa, making it a very important language. This work aims to establish supervised and unsupervised neural machine translation (NMT) baselines between English and Nigerian Pidgin. We implement and compare NMT models with different tokenization methods, creating a solid foundation for future works. 
\end{abstract}

\section{Introduction}

Over 500 languages are spoken in Nigeria, but Nigerian Pidgin is the uniting language in the country. Between three and five million people are estimated to use this language as a first language in performing their daily activities. Nigerian Pidgin is also considered a second language to up to 75 million people in Nigeria, accounting for about half of the country's population according to \cite{bbcpidgin}.

The language is considered an informal lingua franca and offers several benefits to the country. In 2020, 65\% of Nigeria's population is estimated to have access to the internet according to \cite{statista}. However, over 58.4\% of the internet's content is in English language, while Nigerian languages, such as Igbo, Yoruba and Hausa, account for less than 0.1\% of internet content according to \cite{w3tech2020usage}. For Nigerians to truly harness the advantages the internet offers, it is imperative that English content is able to be translated to Nigerian languages, and vice versa. 

This work is a first attempt towards using contemporary neural machine translation (NMT) techniques to perform machine translation for Nigerian Pidgin, establishing solid baselines that will ease  and spur future work. We evaluate the performance of supervised and unsupervised neural machine translation models using word-level and the subword-level tokenization of \cite{sennrich2015neural}. 

\section{Related Work}
\label{related_work}
Some work has been done on developing neural machine translation baselines for African languages. \cite{abbott2018towards} implemented a transformer model which significantly outperformed existing statistical machine translation architectures from English to South-African Setswana. Also, \cite{martinus2019focus} went further, to train neural machine translation models from English to five South African languages using two different architectures - convolutional sequence-to-sequence and transformer. Their results showed that neural machine translation models are very promising for African languages. 

The only known natural language processing work done on any variant of Pidgin English is by \cite{ogueji2019pidginunmt}. The authors provided the largest known Nigerian Pidgin English corpus and trained the first ever translation models between both languages via unsupervised neural machine translation due to the absence of parallel training data at the time.  

\section{Methodology}
\label{methodology}
All baseline models were trained using the Transformer architecture of \cite{vaswani2017attention}. We experiment with both word-level and Byte Pair Encoding (BPE) subword-level tokenization methods for the supervised models. We learned 4000 byte pair encoding tokens, following the findings of \cite{martinus2019focus}. For the unuspervised model, we experiment with only word-level tokenization.

\subsection{Dataset}
The dataset used for the supervised was obtained from the JW300 large-scale, parallel corpus for Machine Translation (MT) by \cite{agic-vulic-2019-jw300}. The train set contained 20214 sentence pairs, while the validation contained 1000 sentence pairs. Both the supervised and unsupervised models were evaluated on a test set of 2101 sentences preprocessed by the \textit{Masakhane}\footnote{{\url{https://github.com/masakhane-io/masakhane/tree/master/jw300_utils/test/}}} group. The model with the highest test BLEU score is selected as the best. 

\subsection{Models}
Unsupervised model training followed \cite{ogueji2019pidginunmt} which used a Transformer of 4 encoder and 4 decoder layers with 10 attention heads. Embedding dimension was set to 300. 

Supervised model training was performed with the open-source machine translation toolkit JoeyNMT by \cite{joey2019}. For the byte pair encoding, embedding dimension was set to 256, while the embedding dimension was set to 300 for the word-level tokenization. The Transformer used for the byte pair encoding model had 6 encoder and 6 decoder layers, with 4 attention heads. For word-level, the encoder and decoder each had 4 layers with 10 attention heads for fair comparison to the unsupervised model. The models were each trained for 200 epochs on an Amazon EC2 p3.2xlarge instance.

\section{Results}
\label{results}

\subsection{Quantitative}

\textbf{English to Pidgin:}

\begin{table}[hbt]
\caption{BLEU Scores (English to Pidgin)}
\label{bleu}
\begin{center}
\begin{tabular}{ll}
\multicolumn{1}{c}{\bf Test BLEU score}  &\multicolumn{1}{c}{\bf Model}
\\ \hline \\
5.18         &Unsupervised (Word-Level) \\
17.73             &Supervised (Word-Level) \\
\bf 24.29             &Supervised (Byte Pair Encoding) \\
\end{tabular}
\end{center}
\end{table}

\textbf{Pidgin to English:}

\begin{table}[hbt]
\caption{BLEU Scores (Pidgin to English)}
\label{bleu}
\begin{center}
\begin{tabular}{ll}
\multicolumn{1}{c}{\bf Test BLEU score}  &\multicolumn{1}{c}{\bf Model}
\\ \hline \\
   7.93      &Unsupervised (Word-Level) \\
\bf24.67             &Supervised (Word-Level) \\
 13.00            &Supervised (Byte Pair Encoding) \\
\end{tabular}
\end{center}
\end{table}

\label{quantitative}
For the word-level tokenization English to Pidgin models, the supervised model outperforms the unsupervised model, achieving a BLEU score of 17.73 in comparison to the BLEU score of 5.18 achieved by the unsupervised model. The supervised model trained with byte pair encoding tokenization outperforms both word-level tokenization models, achieving a BLEU score of 24.29. 

Taking a look at the results from the word-level tokenization Pidgin to English models, the supervised model outperforms the unsupervised model, achieving a BLEU score of 24.67 in comparison to the BLEU score of 7.93 achieved by the unsupervised model. The supervised model trained with byte pair encoding tokenization achieved a BLEU score of 13.00. One thing that is worthy of note is that word-level tokenization methods seem to perform better on Pidgin to English translation models, in comparison to English to Pidgin translation models. 

\subsection{Qualitative}
\label{qualitative}
When analyzed by L1 speakers, the translation qualities were rated very well. In particular, the unsupervised model makes many translations that did not exactly match the reference translation, but conveyed the same meaning. More analysis and translation examples are in the Appendix.

\section{Conclusion}
There is an increasing need to use neural machine translation techniques for African languages. Due to the low-resourced nature of these languages, these techniques can help build useful translation models that could hopefully help with the preservation and discoverability of these languages. 

Future works include establishing qualitative metrics and the use of pre-trained models to bolster these translation models. 

Code, data, trained models and result translations are available here - \href{https://github.com/orevaoghene/pidgin-baseline}{https://github.com/orevaoghene/pidgin-baseline}

\subsubsection*{Acknowledgments}
Special thanks to the \href{masakhane.io}{Masakhane group} for catalysing this work.

\bibliography{iclr2020_conference}
\bibliographystyle{iclr2020_conference}

\newpage
\appendix
\section{Appendix}

\subsection{English to Pidgin translations}

\textbf{Unsupervised (Word-Level):}
\begin{table}[!h]
\centering
\begin{tabular}{ c|c }
\hline
Source      & How has holy spirit helped the Governing Body ?\\ 
Reference      & How holy spirit don take help Governing Body ?\\ 
Model Translation      & ibrahim c how dey do word wey dey guide the governing body . \\
\hline
Source      & What can we learn from Jesus ’ counsel ?\\ 
Reference      & Wetin we fit learn from this advice ? \\ 
Model Translation      & wetin we don learn from jesus ’ counsel \\
\hline
Source      & One student began coming to the kingdom hall .\\
Reference      & One of my student come start to come kingdom hall .\\ 
Model Translation      & one student wey begin dey come di kingdom hall .\\ 
\hline
\end{tabular}
\caption {Unsupervised (Word-Level) Results from English to Nigerian Pidgin}
\end{table}

\textbf{Supervised (Word-Level):}
\begin{table}[!h]
\centering
\begin{tabular}{ c|c }
\hline
Source      & How has holy spirit helped the Governing Body ?\\ 
Reference      & How holy spirit don take help Governing Body ?\\
Model Translation      & How holy spirit take help Governing Body ? \\
\hline
Source      & What can we learn from Jesus ’ counsel ?\\ 
Reference      & Wetin we fit learn from this advice ? \\ 
Model Translation      & Wetin we fit learn from Jesus example ? \\
\hline
Source      & One student began coming to the kingdom hall .\\
Reference      & One of my student come start to come kingdom hall .\\ 
Model Translation      & One day , e start to Kingdom Hall .\\ 
\hline
\end{tabular}
\caption {Supervised (Word-Level) Results from English to Nigerian Pidgin}
\end{table}

\textbf{Supervised (Byte Pair Encoding):}
\begin{table}[!h]
\centering
\begin{tabular}{ c|c }

\hline
Source      & How has holy spirit helped the Governing Body ?\\ 
Reference      & How holy spirit don take help Governing Body ?\\
Model Translation      & How holy spirit take help Governing Body ? \\
\hline
Source      & What can we learn from Jesus ’ counsel ?\\ 
Reference      & Wetin we fit learn from this advice ?  \\ 
Model Translation      & Wetin we fit learn from Jesus example ? \\
\hline
Source      & One student began coming to the kingdom hall .\\
Reference      & One of my student come start to come kingdom hall .\\ 
Model Translation      & One woman come start to dey go meeting .\\ 
\hline
\end{tabular}
\caption {Supervised (Byte Pair Encoding) Results from English to Nigerian Pidgin}
\end{table}

\paragraph{Discussions:} The following insights can be drawn from the example translations shown in the tables above:

\begin{enumerate}
  \item The unsupervised model performed poorly at some simple translation examples, such as the first translation example. 
  \item For all translation models,  the model makes hypothesis that are grammatically and qualitatively correct, but do not exactly match the reference translation, such as the second translation example. 
  \item Surprisingly, the unsupervised model performs better at some relatively simple translation examples than both supervised models. The third example is a typical such case. 
  \item The supervised translation models seem to perform better at longer example translations than the unsupervised example. 
\end{enumerate}

\subsection{Pidgin to English translations}

\textbf{Unsupervised (Word-Level):}
\begin{table}[!h]
\centering
\begin{tabular}{ c|c }
\hline
Source      & How holy spirit don take help Governing Body ?\\ 
Reference    & How has holy spirit helped the Governing Body ? \\
Model Translation      & how holy spirit is to help governing body .\\
\hline
Source      & Wetin we fit learn from this advice ? \\ 
Reference      & What can we learn from Jesus ’ counsel ? \\ 
Model Translation      & what should we learn from this advice ? \\
\hline
\end{tabular}
\caption {Unsupervised (Word-Level) Results from English to Nigerian Pidgin}
\end{table}

\textbf{Supervised (Word-Level):}
\begin{table}[!h]
\centering
\begin{tabular}{ c|c }
\hline
Source      & How holy spirit don take help Governing Body ?\\ 
Reference    & How has holy spirit helped the Governing Body ? \\
Model Translation      & how has holy spirit the governing body ? \\

\hline
Source      & Wetin we fit learn from this advice ? \\ 
Reference      & What can we learn from Jesus ’ counsel ? \\ 
Model Translation      & What can we learn from this ? \\
\hline
\end{tabular}
\caption {Supervised (Word-Level) Results from English to Nigerian Pidgin}
\end{table}

\textbf{Supervised (Byte Pair Encoding):}
\begin{table}[!h]
\centering
\begin{tabular}{ c|c }

\hline
Source      & How holy spirit don take help Governing Body ?\\ 
Reference    & How has holy spirit helped the Governing Body ? \\
Model Translation      & 5 , 6 . ( a ) how did holy spirit help the governing body ? \\
\hline
Source      & Wetin we fit learn from this advice ? \\ 
Reference      & What can we learn from Jesus ’ counsel ? \\ 
Model Translation      & Wtin can we learn from this advice ? \\
\hline
\end{tabular}
\caption {Supervised (Byte Pair Encoding) Results from English to Nigerian Pidgin}
\end{table}

\end{document}